\documentclass{article}
\usepackage{spconf}
\usepackage{amsmath,graphicx}
\usepackage{longtable}
\usepackage{booktabs}
\usepackage{indentfirst}
\usepackage{float}
\usepackage{subfigure}
\usepackage{enumitem}
\usepackage{booktabs}
\usepackage{multirow}

\title{SAR-ShipNet: SAR-Ship Detection Neural Network via Bidirectional Coordinate Attention and Multi-resolution Feature Fusion}
\name{Yuwen Deng \quad Donghai Guan$^{\star}$ \quad Yanyu Chen \quad Weiwei Yuan \quad Jiemin Ji \quad Mingqiang Wei \thanks{This work was supported by the Key Research and Development Program of Jiangsu Province (BE2019012), and Joint Fund of National Natural Science Foundation of China and Civil Aviation Administration of China (U2033202). Corresponding author: D. Guan  (dhguan@nuaa.edu.cn).}}

\address{College of Computer Science and Technology, Nanjing University of Aeronautics and Astronautics\\
Collaborative Innovation Center of Novel Software Technology and Industrialization}
\begin{document}
%
\maketitle
\begin{abstract}
This paper studies a practically meaningful ship detection problem from synthetic aperture radar (SAR) images by the neural network.
We broadly extract different types of SAR image features and raise the intriguing question that whether these extracted features are beneficial to (1) suppress data variations (e.g., complex land-sea backgrounds, scattered noise) of real-world SAR images, and (2) enhance the features of ships that are small objects and have different aspect (length-width) ratios, therefore resulting in the improvement of ship detection.
To answer this question, we propose a SAR-ship detection neural network (call SAR-ShipNet for short), by newly developing Bidirectional Coordinate Attention (BCA) and Multi-resolution Feature Fusion (MRF) based on CenterNet. 
Moreover, considering the varying length-width ratio of arbitrary ships, we adopt elliptical Gaussian probability distribution in CenterNet to improve the performance of base detector models. 
Experimental results on the public SAR-Ship dataset show that our SAR-ShipNet achieves competitive advantages in both speed and accuracy. 
\end{abstract}
\section{Introduction}
\label{sec:intro}
Synthetic Aperture Radar (SAR) is an active microwave imaging sensor with long-distance observation capability in all-day and all-weather conditions and has good adaptability to monitoring the ocean.
In ocean SAR images, ships are the most critical yet small targets to detect when developing a SAR search and tracking system.
SAR-Ship detection aims to find the pre-defined ship objects in a given SAR scene by generating accurate 2D bounding boxes to locate them. 
Although many efforts have been explored to the SAR-ship detection task, it is still not completely and effectively solved, due to the non-trivial SAR imaging mechanism, where various ships are very small and blurred, and even submerged in extremely complicated backgrounds. 
\begin{figure}[t]
\centering
\label{fig:ship}
\subfigure[CenterNet]
{
    \begin{minipage}[b]{.2\linewidth}
        \centering
        \includegraphics[scale=0.21]{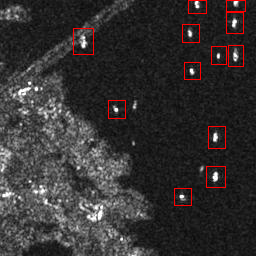} \\
        \includegraphics[scale=0.21]{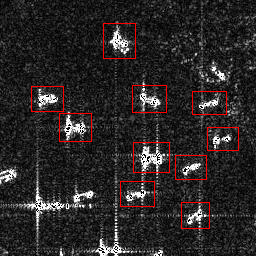}\\
        \includegraphics[scale=0.21]{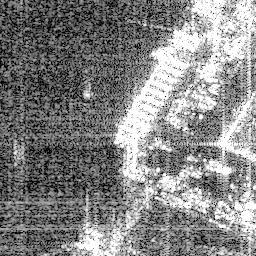}
    \end{minipage}
}
\subfigure[YOLOV4]
{
    \begin{minipage}[b]{.2\linewidth}
        \centering
        \includegraphics[scale=0.21]{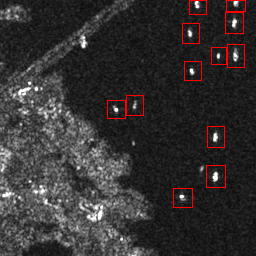}\\
        \includegraphics[scale=0.21]{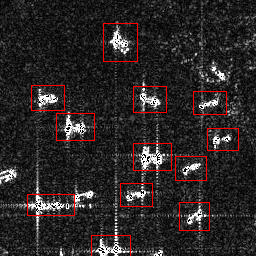}\\
        \includegraphics[scale=0.21]{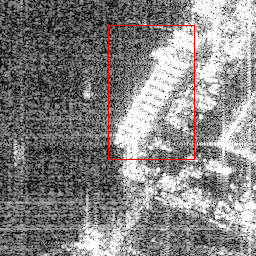}
    \end{minipage}
}
\subfigure[EfficientDet]
{
    \begin{minipage}[b]{.2\linewidth}
        \centering
        \includegraphics[scale=0.21]{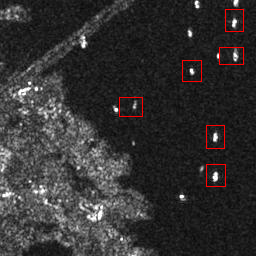}\\
        \includegraphics[scale=0.21]{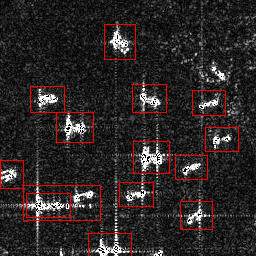}\\
        \includegraphics[scale=0.21]{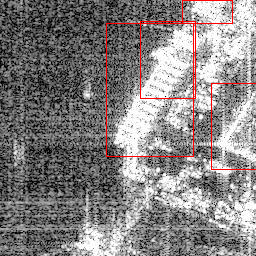}
    \end{minipage}
}
\subfigure[Ours]
{
    \begin{minipage}[b]{.2\linewidth}
        \centering
        \includegraphics[scale=0.21]{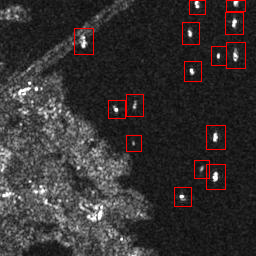}\\
        \includegraphics[scale=0.21]{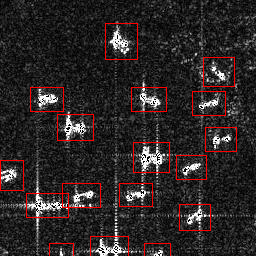}\\
        \includegraphics[scale=0.21]{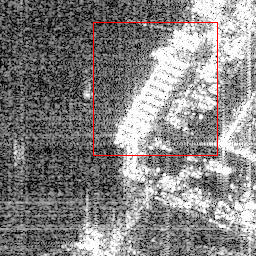}
    \end{minipage}
}
\caption{Ships are often small targets and submerged in extremely complicated backgrounds. Meanwhile, SAR images inevitably contain speckle noise. These adverse factors heavily hinder accurate SAR-Ship detection. When designing a neural network model, it is natural to suppress the extracted features from the adverse factors of surroundings while enhancing the beneficial features from the ship targets. The proposed SAR-ShipNet can deal with the aforementioned problems, therefore leading to better detection results than SOTAs.}
\vspace{-4mm}
\end{figure}

\begin{figure}[t]
\centering  
\subfigure{
\label{Fig.sub.1}
\includegraphics[width=1\linewidth]{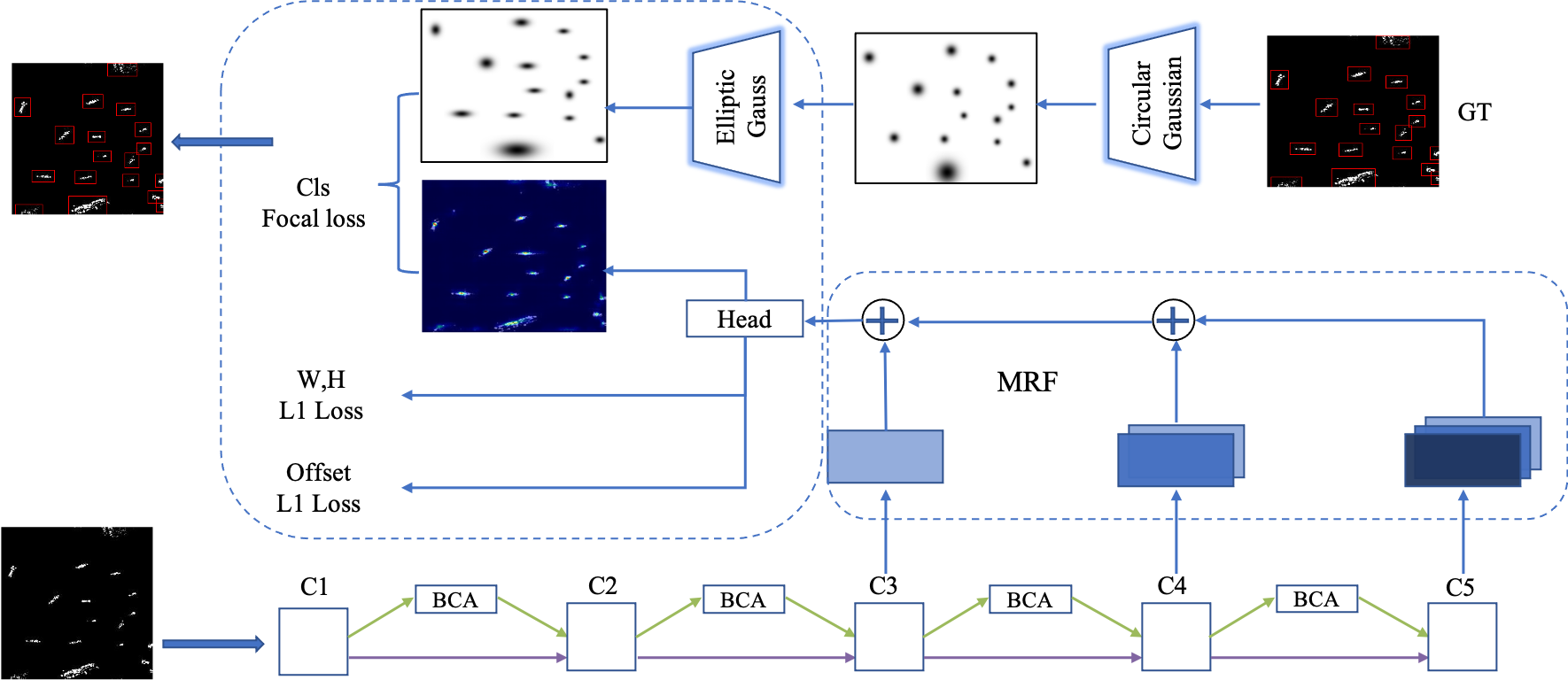}}
\caption{Overview of SAR-ShipNet structure. SAR-ShipNet is composed of three modules: the feature extraction network that the backbone adds the attention mechanism, feature fusion: MRF, and elliptic gauss.}
\label{fig:net}
\vspace{-4mm}
\end{figure}

Traditional SAR target detection methods are mainly based on contrast information, geometric, texture features, and statistics. They are implemented by the hand-crafted feature extractors and classifiers. However, these methods are not only time-consuming but also lead to inaccurate detection results in complicated sea-and-land scenarios. Constant false alarm rate detectors (CFAR)  \cite{gao2008adaptive}, is one of the most commonly used techniques. \cite{gao2016scheme} considers practical application situation and tries to strike a good balance between estimation accuracy and speed. \cite{leng2015bilateral,shi2013ship} introduce a bilateral CFAR algorithm for ship detection and reduced the influence of synthetic aperture radar ambiguity and ocean clutter.

With the development of deep learning, CNN-based detection models have emerged in multitude, which can automatically extract features and get rid of the shortcomings of manually designed features \cite{krizhevsky2012imagenet} for SAR-ship detection. Thus, many researchers begin to use deep learning for SAR ship detection. \cite{cui2019dense} integrates the feature pyramid networks with the convolutional block attention module. \cite{du2019saliency} introduces significant information into the network so that the detector can pay more attention to the object region. \cite{fu2020anchor} proposes an anchor-free network for ship detection, using a balancing pyramid composed of attention-guided and using different levels of features to generate appropriate pyramids. \cite{guo2021CenterNet++} improves CenterNet++ and enhanced ship feature through multi-scale feature fusion and head enhancement. These detectors have achieved great results in SAR-ship detection, there are still many problems with these detectors. These problems include misclassification caused by the high similarity of ships and islands in the complex sea and land scenes, omissions in the detection of small target ships under long-distance satellite observation, and scattering noise in the SAR imaging process. 

Figure~\ref{fig:ship} shows these three types of SAR-ship detection challenges, where the local regions similar to small ship targets
spread over the whole background. Thus, exploring the interaction
information amongst SAR image features in large-range dependencies
to amplify the difference between the ship target and its background is
crucial for robust detection. However, cutting-edge learning models are limited by the locality of CNNs, which behave poorly to capture large-range dependencies.

To solve these challenges, we design a high-speed and effective detector called SAR-ShipNet. We propose a new attention mechanism, i.e., bidirectional coordinated attention (BCA), to solve the effects of complex background noise and islands on ship detection. 
Next, we generate high-resolution feature maps in different feature layers instead of the previous solution of only generating one feature map. This can solve the problem of small ship targets and shallow pixels caused by long-distance detection and scattered noise. 
Finally, considering the change of detection effect caused by the aspect ratio of ships, we adopt an elliptical Gaussian probability distribution scheme to replace the circular Gaussian probability distribution scheme in CenterNet, which significantly improves the detection effect of the detector without any consumption. 
\section{Methodology}
\label{sec:typestyle}
\textbf{Motivation}. 
SAR-ship detection encounters many challenges. Ships in SAR images are small, while backgrounds are usually complex. As a result, the small ship is easily submerged in the complex background, with a low Signal-to-Clutter Ratio (SCR). Besides, the number of ship pixels is much fewer than background pixels. That means the ship and background pixels in an image are of extreme imbalance. Meanwhile, SAR images inevitably contain speckle noise. These factors make SAR-ship detection slightly different from other detection tasks. To develop a high-precision ship detector, one should suppress the extracted features from the adverse factors of backgrounds while enhancing the beneficial features from the ship targets themselves. By completely considering both the adverse and beneficial features of SAR images with ships in them, we broadly extract different types of SAR image features, and 1) suppress data variations (e.g., complex land-sea backgrounds, scattered noise) of SAR images, and 2) enhance the features of ships that are small objects and have different aspect (length-width) ratios, therefore resulting in the improvement of ship detection. We propose a SAR-ship detection neural network (call SAR-ShipNet for short), by newly developing Bidirectional Coordinate Attention (BCA) and Multi-resolution Feature Fusion (MRF) based on CenterNet. SAR-ShipNet is composed of three modules, as shown in Figure~\ref{fig:net}.  
The first module is the feature extraction network that a backbone adds the attention mechanism: BCA. The second module is feature fusion: MRF. The third module is elliptic Gauss.
\subsection{Bidirectional Coordinate Attention}
\begin{figure}[t]
\centering  
\subfigure{
\label{Fig.sub.1}
\includegraphics[width=0.9\linewidth]{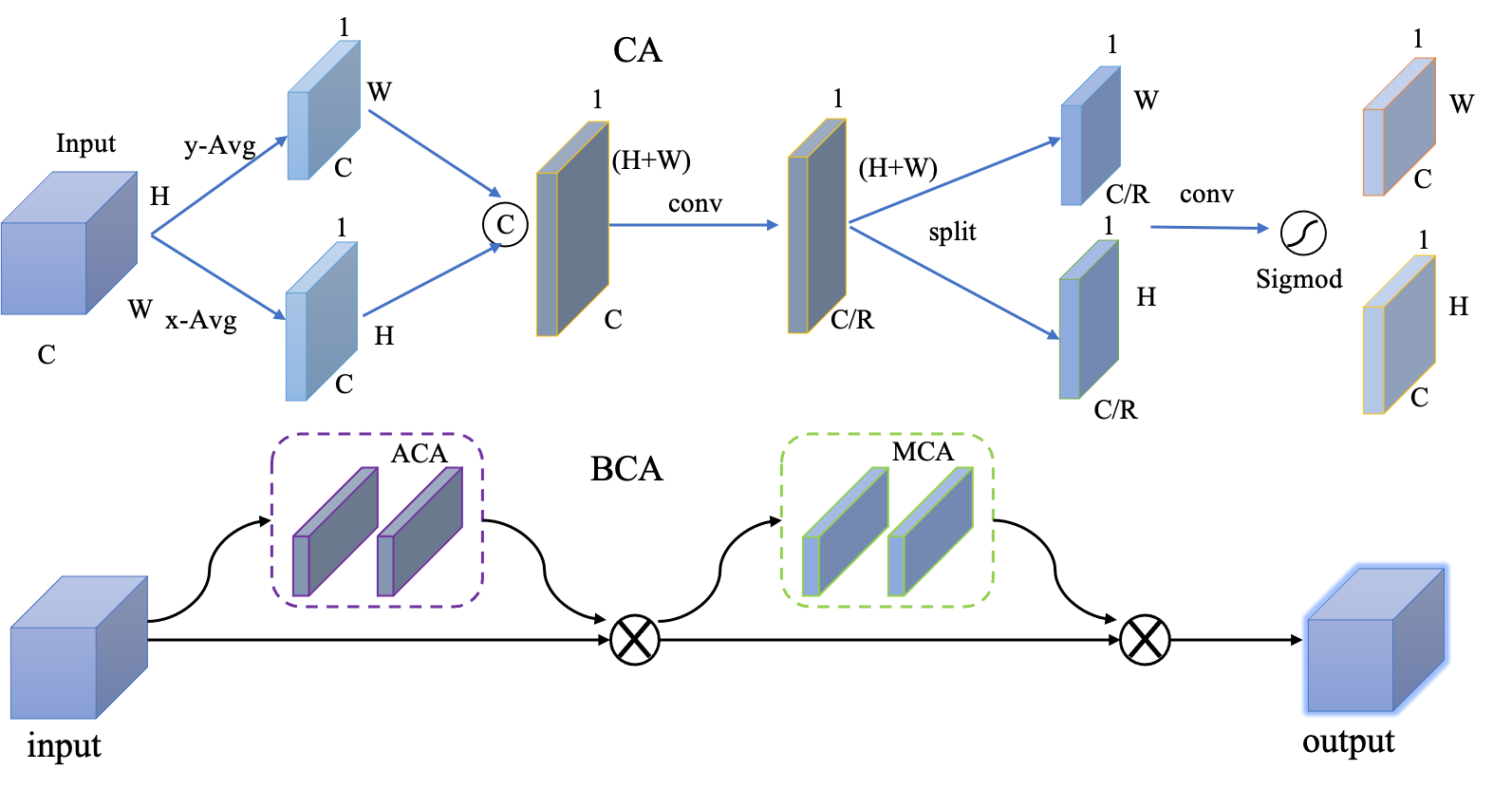}}
\caption{Bidirectional Coordinate Attention mechanism (BCA mechanism). ACA uses Avgpooling to aggregate features, and MCA uses Maxpooling to aggregate features.}
\label{fig:DCA Attention}
\end{figure}

Complicated background islands and other scattered noise affect the effectiveness of ship detection. Inspired by the coordinate attention mechanism (CA) \cite{hou2021coordinate}, we propose a Bidirectional Coordinate Attention mechanism (BCA). CA aggregates information in two directions through Avgpooling and then encodes the generated feature maps into a pair of direction-aware and position-sensitive attention maps, which are complementarily applied to the input feature maps to enhance the representation of the object of interest. But there is a lot of noise redundancy in SAR pictures. Only using average pooling to aggregate information must have noise features to be extracted. It is necessary to ignore unnecessary redundant noise information in SAR pictures. Max pooling information is equally important, thus we propose a BCA mechanism that combines Avg and Max pooling (see Figure~\ref{fig:DCA Attention}).

BCA is formulated as follows:
\begin{equation}
\begin{aligned}
& f_{a}=\delta\left(F_{1}\left[\operatorname{avgpool}\left(x_{c}^{h}\right), \operatorname{avgpool}\left(x_{c}^{w}\right)\right]\right)\\
& g^{h}, g^{w}=\sigma\left(F_{h}\left(f_{a}^{h}\right)\right), \sigma\left(F_{w}\left(f_{a}^{w}\right)\right)\\
& y_{c}(i, j)=x_{c}(i, j) \times g_{c}^{h}(i) \times g_{c}^{w}(j)\\
\end{aligned}
\end{equation}
\begin{equation}
\begin{aligned}
& f_{m}=\delta\left(F_{2}\left[\operatorname{maxpool}\left(y_{c}^{h}\right), \operatorname{maxpool}\left(y_{c}^{w}\right)\right]\right)\\
& z^{h}, z^{w}=\sigma\left(F_{h2}\left(f_{m}^{h}\right)\right), \sigma\left(F_{w2}\left(f_{m}^{w}\right)\right)\\
& \operatorname{output}\left(x_{c}(i, j)\right)=x_{c}(i, j) \times g_{c}^{h}(i)\times g_{c}^{w}(j)\\
&\times z_{c}^{h}(i) \times z_{c}^{w}(j)
\end{aligned}
\end{equation}
where ${x}\in\mathrm{R}^{\mathrm{C}\times\mathrm{W} \times\mathrm{H}}$ is the feature map, $c$ represents the channel index, $avgpool\left(x_{c}^{h}\right)$ and $avgpool\left(x_{c}^{w}\right)$ represents the average pooled output of the c-th channel with height h in the horizontal direction and width w in the vertical direction. $[ ]$ represents the splicing operation of the feature map. ${F}_{1}$ represents the 1$\times$1 convolution. $\delta$ is the non-linear activation function, ${f_{a}}\in\mathrm{R}^{\frac{C}{\mathrm{r}}\times(\mathrm{W}+\mathrm{H}) \times 1} $ is the intermediate feature. $f_{a}^{h} \in \mathrm{R}^{\frac{\mathrm{C}}{\mathrm{r}} \times \mathrm{H} \times 1}$ and $f_{a}^{w} \in \mathrm{R}^{\frac{\mathrm{C}}{\mathrm{r}} \times \mathrm{W} \times 1}$are two vectors obtained by decomposing $f_{a}$, ${F_{h}}$ and ${F_{w}}$ are two 1${\times}$1 convolutions. $\sigma$ is the sigmoid activation function.  $g^{h} \in \mathrm{R}^{\mathrm{C} \times \mathrm{H} \times 1}$ and $g^{w} \in \mathrm{R}^{\mathrm{C} \times \mathrm{W} \times 1}$ are two attention weights respectively. $y_{c}(i, j)$ is the feature point output after avgpooling attention. Similarly, the process of using the maxpooling attention mechanism is consistent with the avgpooling attention mechanism. $\operatorname{output}\left(x_{c}(i, j)\right)$ is the last output of attention through BCA. BCA makes full use of the captured position information through two different information aggregation methods so that the region of interest can be accurately captured. 

\subsection{Multi-resolution Feature Fusion}
The Multi-resolution Feature Fusion module (MRF) is used to enhance the detailed information of small-scale ships to solve the problem of small ship targets and huge differences in surface morphology. In the deep network, if only the last feature layer is used to generate a high-resolution feature map, it is easy to lose the spatial position information of the ship, so we propose an MRF module to enhance ship features.
\begin{figure}[t]
\centering  
\subfigure{
\label{Fig.sub.1}
\includegraphics[width=1\linewidth]{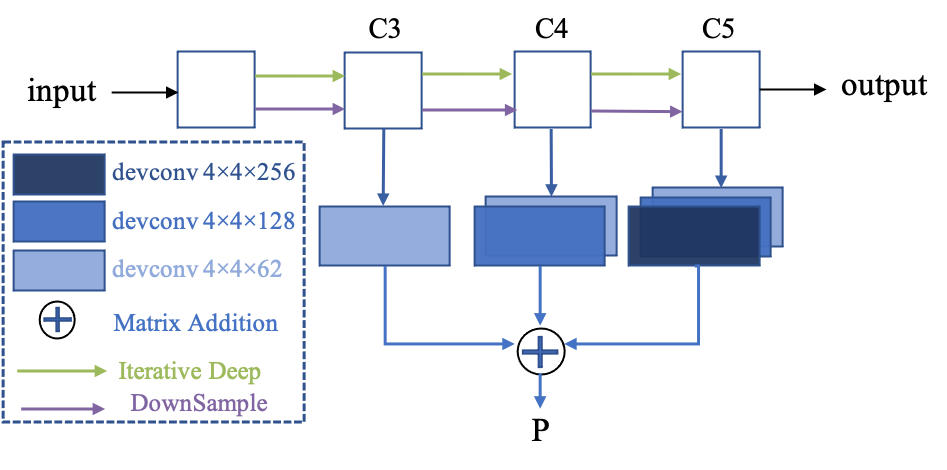}}
\caption{MRF module. {C3, C4, C5} are the feature layers output by the last three stages of Resnet50, Devconv is the deconvolution operation.}
\label{fig:MRF}
\end{figure}
The output of the last three stages of ResNet-50 is defined as \{C3, C4, C5\}. The MRF module uses three feature layers to generate three feature maps of the same size. Figure~\ref{fig:MRF} shows the implementation details of the MRF module. By deconvolution of \{C3, C4, C5\} multiple times. Finally, we merge the three high-resolution feature maps to enhance the detailed features of the ship. The process can be defined as:
\begin{equation}
\small
P=\operatorname{dev}_{-}3\left(C_{5}\right)+\operatorname{dev}_{-}2 \left(C_{4}\right)+\operatorname{dev}_{-}1\left(C_{3}\right)
\end{equation}
where ${dev}_{-}i$ is the deconvolution \cite{2011Adaptive} operation, $i$ is deconvolution times. P is the total feature after fusion. After the feature fusion of the MRF module, it can significantly enhance the feature extraction of ships, reduce the detection interference caused by complex backgrounds, and improve the generalization ability of the model.

\begin{table*}[]
\scriptsize
\centering
\caption{Experimental results of SAR-ShipNet and other different SAR ship detectors.}
\label{fig:all}
\begin{tabular}{ccccccccccccccc}
\toprule
\multirow{2}{*}{Method} &\multirow{2}{*}{Backbone} & \multicolumn{4}{c}{SAR-Ship} & \multicolumn{4}{c}{SSDD} & \multirow{2}{*}{FPS} & \multirow{2}{*}{Parameter}& \multirow{2}{*}{Input-size} \\
\cmidrule(r){3-6} \cmidrule(r){7-10}
&& Precision & Recall & F1& AP0.5
& Precision & Recall & F1& AP0.5 & &&\\
\midrule

YOLOV3 \cite{redmon2018YOLOV3} & DarkNet53 & 92.62 & 70.12 & 80 & 87.24& 90.67 & 67.61 & 77 & 79.06 & 83 & 234MB & 416${\times}$416 \\ 
YOLOV4 \cite{bochkovskiy2020YOLOV4}& DarkNet53 & 94.46 & 70.36 & 81 & 88.76& \textbf{96.94} & 75.65 & 85 & 88.46 & 70 & 245MB & 416${\times}$416 \\
YOLOX \cite{YOLOX2021}& DarkNet53 & 93.65 & 67.51 & 78 & 88.21& 90.78 & 71.36 & 80 & 85.31 & 50 & 97MB & 640${\times}$640 \\
SSD300\cite{liu2016ssd} & VGG16 & 87.79 & 72.48 & 79 & 82.90 & 93.83 & 33.04 & 49 & 74.07 & 142 & 91MB & 300${\times}$300 \\ 
SSD512 \cite{liu2016ssd}& VGG16 & 87.48 &  \textbf{74.58} & 81 & 84.42& 90.07 & 55.22 & 68 & 70.04 & 80 & 92MB & 512${\times}$512 \\ 
RetinaNet\cite{lin2017focal} & ResNet50 & 91.52 & 73.24 & 81 & 88.37& 39.34 & 51.74 & 45 & 37.53 & 49 & 145MB & 600${\times}$600 \\ 
CenterNet\cite{zhou2019objects} & ResNet50 & 94.66 & 60.02 & 74 & 87.44& 92.57 & 59.57 & 72 & 78.86 & 127 & 124MB & 512${\times}$512 \\ 
FR-CNN \cite{ren2015faster} & ResNet50 & 75.68 & 70.95 & 73 & 75.19& 67.51 & 75 & 71 & 71.9 & 15 & 108MB & 600${\times}$600 \\ 
EfficientDet\cite{tan2020efficientdet} & EfficientNet & 89.48 & 71.77 & 80 & 85.20 & 94.33 & 39.78 & 56 & 68.27& 45 & 15MB & 512${\times}$512 \\ 
SAR-ShipNet(ours) & ResNet50 & \textbf{94.85} & 71.31 &  \textbf{81} &  \textbf{90.20}& 95.12 & \textbf{76.30} & \textbf{85} & \textbf{89.08} & 82 & 134MB & 512${\times}$512 \\ 
\bottomrule
\end{tabular}
\vspace{-5mm}
\end{table*}

\subsection{Elliptic Gauss}
In the original CenterNet, the center point of the object needs to be mapped to the heatmap to form a circular Gaussian distribution. This distribution is used to measure the discrete distribution of the center point. For each GT, the key point $p \in R^{2}$ corresponding to category c, then calculate the key points after down sampling $\tilde{p}=\left\lfloor\frac{p}{R}\right\rfloor$. CenterNet splat all ground truth keypoints onto a heatmap $Y \in[0,1]^{\frac{W}{R} \times \frac{H}{R} \times C}$ using a Gaussian kernel $Y_{x y c}=\exp \left(-\frac{\left(x-\tilde{p}_{x}\right)^{2}+\left(y-\tilde{p}_{y}\right)^{2}}{2 \sigma_{p}^{2}}\right)$, where $\sigma_{p}$ is an object size-adaptive standard deviation.
The Gaussian kernel generated by this method is a circular distribution. The parameter $\sigma_{p}$ in the Gaussian kernel is only related to the area of GT, and the aspect ratio of GT is not fully considered. Ships in real life usually have a large aspect ratio. To fully consider the aspect ratio of GT, we are inspired by the elliptic Gaussian method in TtfNet \cite{2019Training}. When the key point $\tilde{p}=\left\lfloor\frac{p}{R}\right\rfloor$ is dispersed on the heatmap, the 2D Gaussian kernel $Y_{x y c}$ is:
\begin{equation}
\small
\mathbf{Y}_{x, y, c}=\exp \left(-\frac{\left(x-\tilde{p}_{x}\right)^{2}}{2 \sigma_{x}^{2}}-\frac{\left(y-\tilde{p}_{y}\right)^{2}}{2 \sigma_{y}^{2}}\right)
\end{equation}
where $\sigma_{x}=\frac{\alpha w}{6}$, $\sigma_{y}=\frac{\alpha h}{6}$, $\alpha$ is a super parameter, $w$ and $h$ are the width and height of GT respectively.

\subsection{Loss Function}
Our training loss function consists of three parts:
\begin{equation}
\small
\begin{aligned}
\operatorname{Loss}=
&\frac{1}{N_{p o s}} \sum_{x y c} F L\left(\hat{p}, p\right)+\frac{\lambda_{1}}{N_{p o s}} \sum_{i}L_{1}\left(\hat{L}_{wh}, L_{w h}\right)\\
& +\frac{\lambda_{2}}{N_{p o s}} \sum_{i} L_{1}(\hat{s}, s)
\end{aligned}
\end{equation}
where ${\hat{p}}$ is the confidence of classification prediction, ${{p}}$ is the ground-truth category label, $FL$ is Focal loss\cite{lin2017focal}. ${ \hat{L}_{wh}}$ are the width and height of the predicted bounding box, ${ L_{w h}}$ are the width and height of the ground-truth bounding box. ${s}$ is the offset $\left(\sigma x_{i}, \sigma y_{i}\right)$ generated by the center point $\left(x_{i}, y_{i}\right)$ of the down-sampling process, ${\hat{s}}$ is the offset predicted value. ${N_{p o s}}$ is the number of positive samples, ${\lambda_{1}}$ and ${\lambda_{2}}$ are the weight parameters. We set  $\lambda_{1}=0.1$ and $\lambda_{2}=1$.

\begin{table}
\scriptsize
\centering
\caption{Ablation experiments on the SAR-Ship dataset.} 
\label{fig:comtab3}
\begin{tabular}{cccc|cccc}
\toprule
\multicolumn{2}{c}{CenterNet}&\multirow{2}{*}{MRF}&\multirow{2}{*}{EGS}& \multirow{2}{*}{Precision}&\multirow{2}{*}{Recall}&\multirow{2}{*}{ F1 }&\multirow{2}{*}{AP0.5}\\
\cline{1-2}
 CA & BCA & & & & &\\
\hline
 $\times$& $\times$& $\times$&$\times$& 94.66& 60.20& 74& 87.44\\
  $\surd$& $\times$& $\times$&$\times$& 96.95& 51.80& 68& 88.56\\
   $\times$& $\surd$& $\times$&$\times$& 96.76& 57.71& 72& 89.10\\
    $\times$& $\surd$& $\surd$&$\times$& \textbf{97.06}& 50.19& 66& 89.40\\
    $\times$& $\surd$& $\surd$&$\surd$& 94.85& \textbf{71.31}& \textbf{81}& \textbf{90.20}\\

\toprule
\end{tabular}
\vspace{-5mm}
\end{table}

\section{Experiments}

\subsection{Experimental Dataset}
\label{sec:majhead}
We directly evaluate the SAR-ShipNet model on the SAR-Ship \cite{wang2019sar} and SSDD\cite{li2017ship} dataset. The SAR-ship dataset contains ship slices (43819) and the number of ships (59535) and the size of the all ship slices is fixed at 256${\times}$256 pixels. The SSDD data set has a total of 1160 images and 2456 ships. We randomly divide the data set into the training set, validation set, and test set at a ratio of 7:1:2.

\subsection{Experimental results}
\begin{table}
\scriptsize
\centering
\caption{SAR-ShipNet test results of different $\alpha$.}
\begin{tabular}{c| c | c | c | c}
\hline
		Parameter  & Precision & Recall & F1 & AP0.5 	\\
		\hline
	
	    $\alpha$ = 0.1           & 96.16  			& 6.12  				& 12  & 87.40        \\
	$\alpha$ = 0.2   		    & 97.84		    & 7.2   			    & 13  & 88.16       \\
	$\alpha$ = 0.3       			& \textbf{98.12}  			& 24.38    			    & 39  & 88.81      \\
	$\alpha$ = 0.4        			& 97.58  			& 42.11    			    & 59  & 89.48      \\
	$\alpha$ = 0.5        			& 95.86  			& 63.12    			    & 76  & 89.80      \\$\alpha$ = 0.6        			& 97.36  			& 55.82    			    & 71  & 90.03      \\
	$\alpha$ = 0.7        			& 96.61  			& 61.58    			    & 75  & 89.85      \\
	$\alpha$ = 0.8        			& 94.85 			& \textbf{71.31}    			    & \textbf{81}  & \textbf{90.20}      \\

	$\alpha$ = 0.9       			& 95.88  			& 65.2    			    & 78  & 90.16
	\\
		\hline
	 Circular Gaussian    			& 97.06 	 		& 50.19  				& 66  & 89.40        \\
	\end{tabular}
	\label{tab:e}
\vspace{-5mm}
\end{table}

 We evaluate our SAR-ShipNet on 4 evaluation metrics and compare it with other methods. Table~\ref{fig:all} shows the quantitative results of all the methods in two datasets. Compared with other detectors SAR-ShipNet achieves the best $F_{1}$, and AP on two datasets, indicating that our model has the best overall performance. This is because SAR-ShipNet uses the attention mechanism to pay more attention to ship features, and uses feature fusion to strengthen small targets and fully consider the aspect ratio of the ship. Experiments show that our model can achieve the best comprehensive performance on both the large dataset SARShip and the small dataset SSDD. Table~\ref{fig:comtab3} shows the ablation experimental results. It can be found that CA, BCA, MRF, and elliptic Gauss can increase the detection performance of the model. In particular, after adding the attention mechanism, the precision and AP have been improved, which shows that our model reduces the misclassification of islands and backgrounds into ships. Table~\ref{tab:e} shows the experimental results of the effect of hyperparameter $\alpha$ on SAR-Ship dataset. When $\alpha$ = 0.8, we get the best AP (90.20).
 
\section{Conclusion}
In this paper, we propose an effective SAR-ShipNet for SAR-ship detection. SAR-ShipNet mainly has three modules: the BCA mechanism, the MRF module, and the elliptic Gaussian module. BCA mechanism is used to solve the problem of ship detection in complex backgrounds. It can make the model pay attention to ship features as much as possible while ignoring background noise. The MRF module is used to solve the problems of small ship sizes and shallower pixels in long-distance observation. Elliptical Gauss fully considers the influence of ship aspect ratio detection. Experimental results show that our SAR-ShipNet achieves a competitive detection performance.
\bibliographystyle{IEEE}
\bibliography{ref}

\end{document}